\documentclass[10pt,twocolumn,letterpaper]{article}

\usepackage{cvpr}
\usepackage{times}
\usepackage{epsfig}
\usepackage{graphicx}
\usepackage{amsmath}
\usepackage{amssymb}

\usepackage{makecell}
\usepackage{multirow}
\usepackage{threeparttable}
\usepackage{bbding}
\usepackage{booktabs} % To thicken table lines
\usepackage{url}
\usepackage{array}
\usepackage{color}

\def\ie{{\em i.e.}}
\def\etal{{\em et al.}}

\usepackage[breaklinks=true,bookmarks=false]{hyperref}

\cvprfinalcopy % *** Uncomment this line for the final submission

\def\cvprPaperID{****} % *** Enter the CVPR Paper ID here

\begin{document}

%%%%%%%%% TITLE
\title{Static and Dynamic Fusion for Multi-modal Cross-ethnicity \\ Face Anti-spoofing}

\author{Ajian Liu$^{\rm 1*}$, Zichang Tan$^{\rm 2}$\thanks{Equal Contribution}, Xuan Li$^{\rm 3}$, Jun Wan$^{\rm 2}$\thanks{Corresponding Author, email:  jun.wan@ia.ac.cn},\ \ Sergio Escalera$^{\rm 4}$, Guodong Guo$^{\rm 5}$,  Stan Z. Li$^{\rm 1,2}$\vspace{1.2mm}\\
	{$^{\rm 1}$MUST, Macau, China;
		$^{\rm 2}$NLPR, CASIA, China;}
	{$^{\rm 3}$BJTU, China;
		$^{\rm 4}$CVC, UB, Spain;
		$^{\rm 5}$Baidu Research, China}\\
	{\tt\small
		%ajianliu92@gmail.com\vspace{-1.2mm},\{tanzichang2016,jun.wan,szli\}@ia.ac.cn
		\{jun.wan, szli\}@ia.ac.cn, sergio@maia.ub.es,guoguodong01@baidu.com
	}\\
	{\tt\small
		%18120387@bjtu.edu.cn,sergio@maia.ub.es,guoguodong01@baidu.com
	}\\
}

\maketitle
%\thispagestyle{empty}

%%%%%%%%% ABSTRACT
\begin{abstract}
   Regardless of the usage of deep learning and handcrafted methods, the dynamic information from videos and the effect of cross-ethnicity are rarely considered in face anti-spoofing. In this work, we propose a static-dynamic fusion mechanism for multi-modal face anti-spoofing. Inspired by motion divergences between real and fake faces, we incorporate the dynamic image calculated by rank pooling with static information into a conventional neural network (CNN) for each modality (i.e., RGB, Depth and infrared (IR)). Then, we develop a partially shared fusion method to learn complementary information from multiple modalities. Furthermore, in order to study the generalization capability of the proposal in terms of cross-ethnicity attacks and unknown spoofs, we introduce the largest public cross-ethnicity Face Anti-spoofing  (CASIA-SURF CeFA) dataset, covering 3 ethnicities, 3 modalities, 1607 subjects, and 2D plus 3D attack types. Experiments demonstrate that the proposed method achieves state-of-the-art results on CASIA-SURF CeFA, CASIA-SURF, OULU-NPU and SiW.
\end{abstract}

%%%%%%%%% BODY TEXT
\section{Introduction}
%按照摘要的优缺点，进行详细说明，然后引出我们的创新点：方法，数据
%创新点： 1）staic_dynamic 方法（单模态）； 2) partial shared 网络 融合；3）数据库；4）实验最好
%Face anti-spoofing is a key element to avoid security breaches in face recognition systems~\cite{Boulkenafet2016Face,Boulkenafet2017Face,Liu2018Learning,shao2019multi}. It has been widely applied in financial payment, access control, phone unlock and surveillance, just to mention a few.

\begin{figure}[t]
	\begin{center}
		\includegraphics[width=1.0\linewidth]{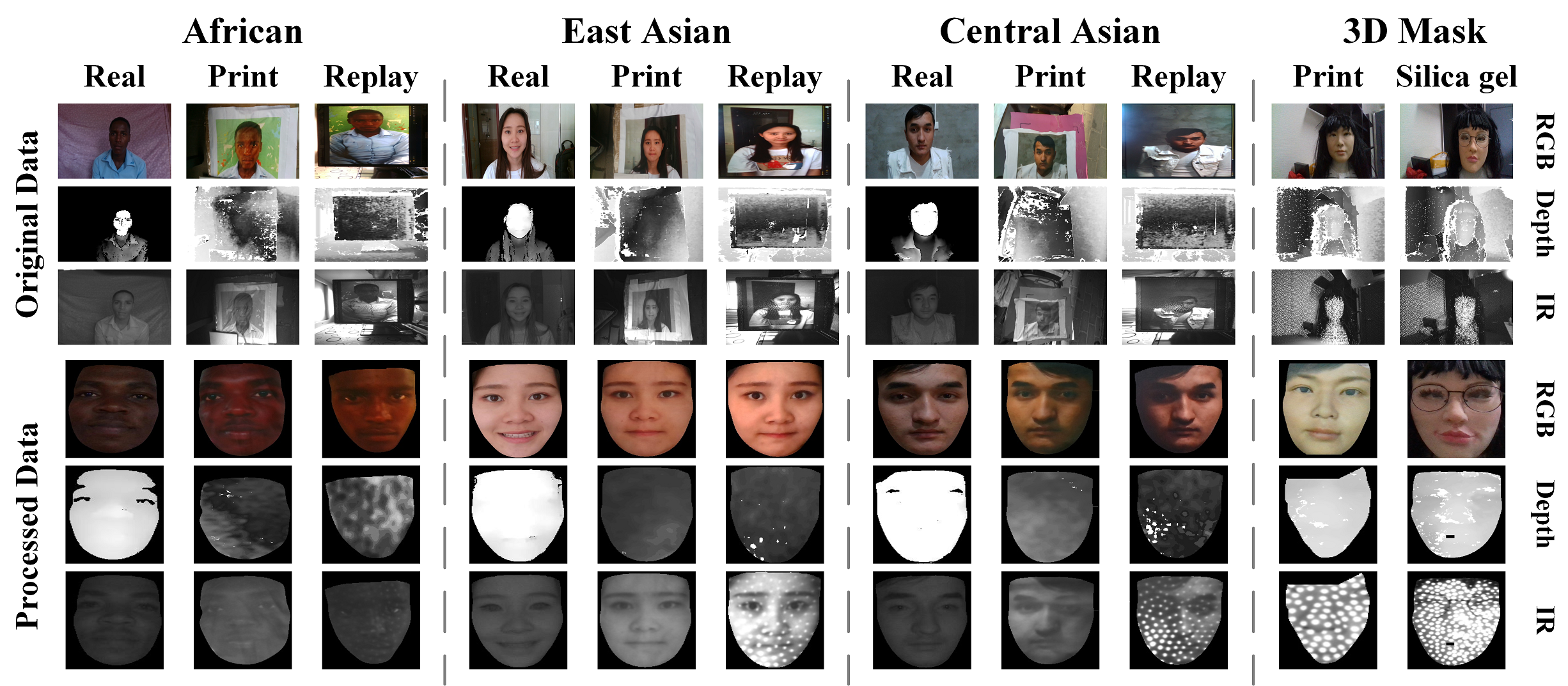}
	\end{center}
	\caption{Samples of the CASIA-SURF CeFA dataset. It contains 1607 subjects, 3 different ethnicities (\ie, Africa, East Asia, and Central Asia), with 4 attack types (\ie, print attack, replay attack, 3D print and silica gel attacks).}
	\label{samp_data}
	% \label{fig:onecol}
\end{figure}
In order to enhance security of face recognition systems, the presentation attack detection (PAD) technique is a vital stage prior to visual face recognition~\cite{Boulkenafet2016Face,Boulkenafet2017Face,Liu2018Learning,shao2019multi}. %Therefore, face anti-spoofing is vital step to ensure the security of face recognition systems.
Most works in face anti-spoofing focus on still-images, including RGB, Depth or IR). These methods can be divided into two main categories: handcrafted methods~\cite{Boulkenafet2016Face,de2013can,patel2016secure} and deep learning based methods~\cite{feng2016integration,li2016original,patel2016cross}. Handcrafted methods attempt to extract texture information or statistical features (\ie, HOG~\cite{Komulainen2014Context,yang2013face} and LBP~\cite{maatta2011face,de2013can}) to distinguish between real and spoof faces. Deep learning based methods automatically learn discriminative features from input images for face anti-spoofings~\cite{li2016original,patel2016cross}.
However, the analysis of motion divergences between real and fake faces received little attention. In order to improve robustness in real applications, some temporal-based methods~\cite{Pan2007Eyeblink,patel2016cross} have been proposed, which require from a constrained human guided interaction, such as movements of of eyes, lips, and head. However, this does not provide with a natural user friendly interaction. %Even more importantly, these methods~\cite{Pan2007Eyeblink,patel2016cross} could become vulnerable if someone presents a replay attack or a print photo attack with cut eye/mouth regions.
%Inspired by the motion divergences between real and fake faces, we propose a static- and dynamic-based network (SD-Net) to learn static-dynamic features for each modality (\ie, RGB, depth, IR) without any human guide interaction. %We incorporate the dynamic image calculated by rank pooling~\cite{fernando2017rank} with static information into the SD-Net for each modality (\ie, RGB, depth, IR)
Different from those works, we capture the temporal/dynamic information by using dynamic image generated by rank pooling~\cite{fernando2017rank}, which doesn't need any human guide interaction. Moreover, inspired by the motion divergences between real and fake faces, a static- and dynamic-based network (SD-Net) is further formulated by taking the static and dynamic images as the input.
%\textcolor{blue}{Different from those works, we capture the temporal/dynamic information by using the dynamic image generated by rank pooling~\cite{fernando2017rank}, which doesn't need any human guide interaction. Moreover, to learn both static and dynamic features, a static- and dynamic-based network (SD-Net) is further formulated by taking the static and dynamic images as the input.}

Multi-modal face anti-spoofing have also absorbed an increasing number of researchers in recent two years. Some fusion methods~\cite{DBLP:conf/cvpr/abs-1812-00408,parkin2019recognizing} are published, which restrict the interactions among different modalities since they are independent before the fusion. But it
is difficult for different modalities to effectively utilize the modality relatedness from the beginning of the network to its end to boost the overall performance. %In order to capture correlations and complementary semantics among different modalities, a partially shared branch multi-modality network (PSMM-Net) are designed, which incorporates SD-Nets and a shared network. %This is done in two different ways. The first is forward feeding of fused SD-Net features to the shared branch, and the second is backward feeding from shared branch modules output to SD-Net block inputs.
In this paper, we propose a partially shared branch multi-modal network (PSMM-Net)
with allowing the exchanges and interactions among different modalities, aiming to capture  correlated and complementary features. %More specially, PSMM-Net consists of a shared branch and three SD-Nets, each of which learns the features from one of RGB, depth, IR modalities.

\newcommand{\tabincell}[2]{\begin{tabular}{@{}#1@{}}#2\end{tabular}}
\begin{table}[h]\setlength\tabcolsep{3.5pt}
\begin{center}
\scalebox{0.73}{
\begin{tabular}{|c|c|c|c|c|c|c|c|}
\hline
Dataset & Year & \# sub & \# num & Attacks & Mod. & Dev. & Eth. \\ \hline \hline
Replay-Attack~\cite{Chingovska2012On}& 2012  & 50 & 1200 &Pr,Re & R & CR & *\\ \hline
CASIA-FASD~\cite{Zhang2012A} & 2012 & 50  & 600 &Pr,Cu,Re & R & CR & *\\\hline
3DMAD~\cite{ERDOGMUS_BTAS-2013} & 2014  & 17 & 255  &M& R/D& CR/K  & *\\ \hline
MSU-MFSD~\cite{Wen2015Face}  & 2015 & 35 & 440  &Pr,Re & R & P/L & * \\ \hline
Replay-Mobile~\cite{Costa2016The}  & 2016 & 40  & 1030  & Pr,Re & R & P  & *\\ \hline
Msspoof~\cite{msspoof-2015}  & 2016  & 21 & 4704\textsuperscript{\textit{i}}& \begin{tabular}[c]{@{}c@{}}Pr\end{tabular} & R/I & CR/CI & *\\ \hline
OULU-NPU~\cite{Boulkenafet2017OULU} & 2017 & 55 & 5940 &Pr,Re & R & \begin{tabular}[c]{@{}c@{}}CR\end{tabular} & * \\ \hline
SiW~\cite{Liu2018Learning}  & 2018  & 165  & 4620  & \begin{tabular}[c]{@{}c@{}}Pr,Re\end{tabular} & R& CR  & \tabincell{c}{AS/A/\\U/I}\\ \hline
CASIA-SURF~\cite{DBLP:conf/cvpr/abs-1812-00408}  & 2019  & 1000  & 21000  & \begin{tabular}[c]{@{}c@{}}Pr,Cu\end{tabular}  & R/D/I & S  & E\\ \hline
\multirow{4}{*}{\tabincell{c}{CASIA-SURF CeFA\\(Ours)}} & \multirow{4}{*}{2019} & 1500 & 18000& Pr, Re & \multirow{3}{*}{R/D/I} &\multirow{3}{*}{S} & \multirow{3}{*}{A/E/C} \\ \cline{3-5}
&  & 99 & 5346 &  M &  & & \\ \cline{3-5}
&  & 8 & 192 & G &  & & \\ \cline{3-8}
&  & \multicolumn{6}{c|}{Total: 1607 subjects, 23538 videos} \\ \hline
\end{tabular}
}
\caption{Comparisons among existing face PAD databases. (\textit{i} and * indicates the dataset only contains imges and does not provide specify ethnicities, respectively. Mod.: modalities, Dev.: devices, Eth.: ethnicities, Pr: print attack, Re: replay attack, Cu: Cut, M: 3D print face mask, G: 3D silica gel face mask, R: RGB, D: Depth, I: IR, CR: RGB Camera, CI: IR Camera, K: Kinect, P: Cellphone, L: Laptop, S: Intel Realsense, AS: Asian, A: Africa, U: Caucasian, I: Indian, E: East Asia, C: Central Asia.)}
\label{tab1}
\end{center}
\end{table}

Furthermore, data plays a key role in face anti-spoofing tasks. About existing face anti-spoofing datasets, such as CASIA-FASD~\cite{Zhang2012A}, Replay-Attack~\cite{Chingovska2012On}, OULU-NPU~\cite{Boulkenafet2017OULU}, and SiW~\cite{Liu2018Learning}, the amount of sample is relatively small and most of them just contain the RGB modality. The recently released CASIA-SURF~\cite{DBLP:conf/cvpr/abs-1812-00408} includes 1,000 subjects and RGB, Depth and IR modalities. Although this provides with a larger dataset in comparison to the existing alternatives, it suffers from limited attack types (2D print attack) and single ethnicity (Chinese people). Overall, the effect of cross-ethnicity for face anti-spoofing received little attention in previous works.
Therefore, we introduce CASIA-SURF CeFA dataset, the largest dataset in terms of subjects (see Table~\ref{tab1}). In CASIA-SURF CeFA, attack types are diverse, including printing from cloth, video replay attack, 3D print and silica gel attacks. More importantly, it is the first public dataset designed for exploring the impact of cross-ethnicity in the study of face anti-spoofing. Some samples of the CASIA-SURF CeFA dataset are shown in Fig.~\ref{samp_data}.

\iffalse
In order to address previous issues, in this work we first propose a static-dynamic fusion mechanism for face anti-spoofing, which is inspired by motion divergences between real and fake faces. We incorporate the dynamic image calculated by rank pooling~\cite{fernando2017rank} with static information into a CNN for each modality (\ie, RGB, Depth, IR) and design a partially shared fusion method to learn complementary information from the multiple modalities. Second, we introduce CASIA-SURF CeFA, the largest dataset in terms of subjects, from 1,000 to 1,607 compared with the CASIA-SURF~\cite{DBLP:conf/cvpr/abs-1812-00408} dataset. In CASIA-SURF CeFA, attack types are diverse, including printing from cloth, video replay attack, 3D print and silica gel attacks. More importantly, it is the first public dataset designed for exploring the impact of cross-ethnicity in the study of face anti-spoofing. Some samples of the CASIA-SURF CeFA dataset are shown in Fig.~\ref{samp_data}.
\fi

To sum up, the contributions of this paper are summarized as follows: (1) We propose the SD-Net to learn both static and dynamic features for single modality . It is the first work incorporating dynamic images for face anti-spoofing. (2) We propose the PSMM-Net to learn complementary information from multi-modal data in videos. (3) We release the CASIA-SURF CeFA dataset, which includes $3$ ethnicities, $1607$ subjects and $4$ diverse 2D/3D attack types. (4) Extensive experiments of the proposed method on CASIA-SURF CeFA and other 3 public datasets verify its high generalization capability.

\iffalse
To sum up, the contributions of this paper are summarized as follows: (1) We propose a static- and dynamic-based network (SD-Net) for single modality to learn both static and dynamic features. It is the first work incorporating dynamic images for face anti-spoofing. (2) We propose a partially shared multi-modality network (PSMM-Net) to learn complementary information from multi-modal data in videos. (3) We release the CASIA-SURF CeFA dataset, which includes $3$ ethnicities, $1607$ subjects and $4$ diverse 2D/3D attack types. (4) Extensive experiments of the proposed method on proposed CASIA-SURF CeFA and 4 public dataset verify its high generalization capability.
\fi

\iffalse
\begin{itemize}
\item We propose a static- and dynamic-based network (SD-Net) for single-modal to learn hybrid features. It is the first work incorporating dynamic images for face anti-spoofing.
\item We propose a partially shared multi-modal network (PSMM-Net) to learn complementary information from multi-modal data in videos.
\item We release the CASIA-SURF CeFA dataset, which includes 3 ethnicities, 1607 subjects and 4 diverse 2D/3D attack types.
\item Extensive experiments of the proposed method on proposed CASIA-SURF CeFA and existing public dataset verify its significance and generalization capability in both inter- and intra-dataset evaluation.
\end{itemize}
\fi
\section{Related Work}
%Face anti-spoofing has made great progress in recent years and lots of methods have been proposed with the help of face anti-spoofing datasets. In this section, we first review some representative methods and then summarize the existing face anti-spoofing datasets.

 %This section reviews representative methods and datasets in face anti-spoofing.

%分两部分：第一部分 1）基于静态图像信息的人脸防伪；2）基于时序信息的人脸防伪；第二部分 数据库的简单介绍和对比 说明当前数据库的缺点

\subsection{Methods}
\textbf{Image-based Methods.}  Image-based methods take still images as input, \ie, RGB, Depth or IR. Classical approaches based on handcrafted features, such as HOG~\cite{yang2013face}, LBP~\cite{maatta2011face,de2013can}, SIFT~\cite{patel2016secure} or SURF~\cite{Boulkenafet2016Face} together with traditional classifiers, such as SVM or LDA, to perform binary anti-spoofing predictions. However, those methods lack of good generalization capability when testing conditions vary, such as lighting and background. Owing to the success  of deep learning strategies over handcrafted alternatives in computer vision, some works~\cite{feng2016integration,li2016original,patel2016cross} extended feature vectors with features from CNN networks for face anti-spoofing. %The works of~\cite{li2016original,patel2016cross} use fine-tuned features from a pre-trained model and take it to extra features.
Authors of~\cite{atoum2017face,Liu2018Learning} presented a two-stream network using RGB and Depth images as input. The work of~\cite{liu2019deep} proposes a deep tree network to model spoofs by hierarchically learning sub-groups of features. However, previous methods do not consider any kind of temporal information for face anti-spoofing.

\textbf{Temporal-based Methods.} %To increase the robustness of models, some temporal-based methods~\cite{Pan2007Eyeblink,patel2016cross,shao2017deep} focus on the movement of key parts of face (\ie, eye-blinking, lip, head movements). However, those methods are not a friendly and nature interaction way for users. More importantly, those methods~\cite{Pan2007Eyeblink,patel2016cross,shao2017deep} would become vulnerable if someone present a replay attack or a print photo attack with eye or mouth portion being cut.
In order to improve robustness in real applications, some temporal-based methods~\cite{Pan2007Eyeblink,patel2016cross} have been proposed, which require from a constrained human guided interaction, such as movements of of eyes, lips, and head. However, those methods do not provide with a natural user friendly interaction. Even more importantly, these methods~\cite{Pan2007Eyeblink,patel2016cross} could become vulnerable if someone presents a replay attack or a print photo attack with cut eye/mouth regions. Given that the Photoplethysmography (rPPG) signals (\ie{ heart plus signal}) can be detected from real but not spoof, Liu~\etal~\cite{Liu2018Learning} proposed a CNN-RNN model to estimate rPPG signals with sequence-wise supervision and face depth with pixel-wise supervision. The estimated depth and rPPG are fused to distinguish real and fake faces. Feng~\etal~\cite{feng2016integration} distinguished between real and fake samples based on the the difference between image quality and optical flow information. Yang~\etal~\cite{yang2019face} proposed a spatio-temporal attention mechanism to fuse global temporal and local spatial information. All previous methods rely on a single visual modality, and no work considers the effect of cross-ethnicity for anti-spoofing.

\textbf{Multi-modal Fusion Methods.}  Zhang~\etal~\cite{DBLP:conf/cvpr/abs-1812-00408} proposed a fusion network with 3 streams using ResNet-18 as the backbone, where each stream is used to extract low level features from RGB, Depth and IR data, respectively. Then, these features are concatenated and passed to the last two residual blocks. Similar to~\cite{DBLP:conf/cvpr/abs-1812-00408}, Aleksandr~\etal~\cite{parkin2019recognizing} used a fusion network with 3 streams. They used ResNet-34 as the backbone and multi-scale feature fusion at all residual blocks. Tao~\etal~\cite{shen2019facebagnet} proposed a multi-stream CNN architecture called FaceBagNet, which uses patch-level images as input and modality feature erasing (MFE) operation to prevent overfitting and obtain more discriminative fused features. All previous methods just consider as a key fusion component the concatenation of features from multiple modalities. Unlike~\cite{DBLP:conf/cvpr/abs-1812-00408,parkin2019recognizing,shen2019facebagnet}, we propose the PSMM-Net, where three modality-specific networks and one shared network are connected by using a partially shared structure to learn discriminative fused features for face anti-spoofing.

\subsection{Datasets}
Table~\ref{tab1} lists existing face anti-spoofing datasets. One can see that before 2019 the maximum number of available subjects was 165 on the SiW dataset~\cite{Liu2018Learning}. That was clearly limiting the generalization ability of new approaches for cross-dataset evaluation. Most of the datasets just contain RGB data, such as Replay-Attack~\cite{Chingovska2012On}, CASIA-FASD~\cite{Zhang2012A}, SiW~\cite{Liu2018Learning} and OULU-NPU~\cite{Boulkenafet2017OULU}. %However, these methods are sensitive to 3D print and silica gel face mask.
Recently, the CASIA-SURF~\cite{DBLP:conf/cvpr/abs-1812-00408} has been released, including 1000 subjects with three modalities, namely RGB, Depth and IR.
Although this relieved the problem of the amount of data, it is limited in terms of attack types (only 2D print attack) and only includes 1 ethnicity (Chinese people). As shown in Table~\ref{tab1}, most datasets do not provide ethnicity information, except SiW and CASIA-SURF. Although the SiW dataset provides four ethnicities, it still does not consider the effect of cross-ethnicity for face anti-spoofing. This limitation also holds for the CASIA-SURF dataset.

\section{Proposed Method}
\iffalse
In addition to texture information extraction from silent face images, some prior works~\cite{Liu2018Learning,DBLP:conf/cvpr/abs-1812-00408} %~\cite{FASTD2018arionxiv,Liu2018Learning,DBLP:conf/cvpr/abs-1812-00408}
also verified the key role of motion and multi-modal information for face anti-spoofing. How to efficiently extract the motion information in video sequences and fuse the complementary information among multiply modalities are two main concerns in our method.
\fi
%This section first introduces the static- and dynamic-based network (SD-Net) to fuse static-dynamic features for single modality. The SD-Net diagram is shown in Fig.~\ref{fig:static_dynamic}. Then, the partially shared multi-modality network (PSMM-Net) for multi-modal data fusion is presented.

%multi-modality to discover and capture the correlations and complementary features among those modalities.%}  %As shown in Fig.~\ref{fig:modal_fusion}, this model is composed of $3$ branches, which correspond to the $3$ modalities (\ie, RGB, Depth and IR) of the input network face image respectively. Each branch considers the static texture and dynamic motion information of each modality image simultaneously and the different modality information of the three branches is dense fusion.

\begin{figure}[t]
\centering
\setlength{\tabcolsep}{8pt}
\includegraphics[width=0.95\linewidth]{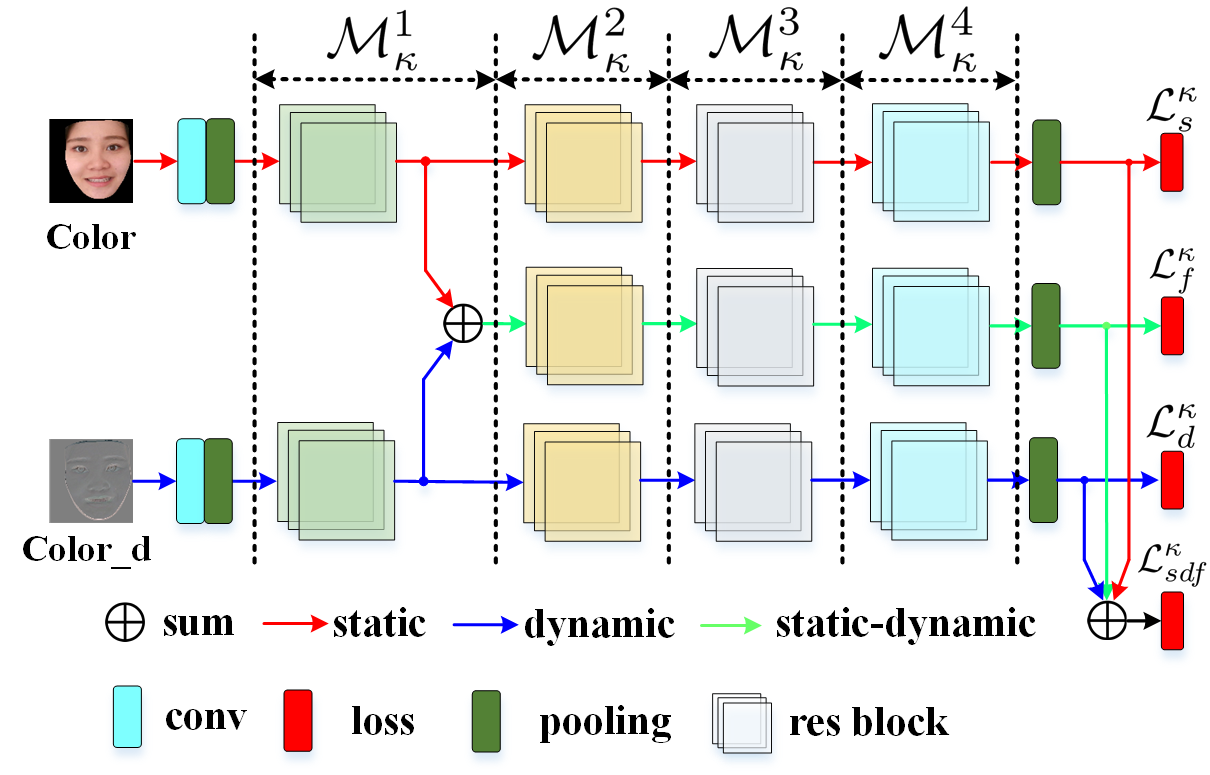}
\caption{SD-Net diagram, showing a single-modal static-dynamic-based network.
We take the RGB modality and its  corresponding dynamic image as an example.
%We use ResNet-18~\cite{he2016deep} backbone.
This architecture includes three branches: static (red arrow), dynamic (blue arrow) and static-dynamic (green arrow). The static-dynamic branch fuses the static and dynamic features of first res block outputs from static and dynamic branches (best viewed in color).
%Taking the RGB modality as an example, inputs are the RGB image and its corresponding dynamic image.
}
\label{fig:static_dynamic}
\end{figure}

\subsection{SD-Net for Single-modal}
{\flushleft \textbf{Single-modal Dynamic Image Construction.}} Rank pooling~\cite{fernando2017rank,wang2018cooperative} defines a rank function that encodes a video into a feature vector. The learning process can be seen as a convex optimization problem using the RankSVM~\cite{smola2004tutorial} formulation in Eq.\ref{Eq:ranksvm}. Let RGB (Depth or IR) video sequence with $K$ frames be represented as $<{\bf I}_1, {\bf I}_2, ..., {\bf I}_i,...,{\bf I}_K>$,  and ${\bf I}_i$ denotes the average of RGB (Depth or IR) features over time up to $i$-frame. The process is formulated below.
\begin{equation}\label{Eq:ranksvm}
\begin{split}
\underset{{\bf d}}{argmin}
 & \frac{1}{2} \| {\bf d} \|^2 +  \delta \times \sum_{i>j}{\xi_{ij}}  \\
 s.t. \ {\bf d}^T \cdot & ({\bf I}_i-{\bf I}_j) \geq 1- \xi_{ij}, \ \xi_{ij} \geq 0
\end{split}
\end{equation}
where $\xi_{ij}$ is the slack variable, and $\delta=\frac{2}{K(K-1)}$.%, $d$ is the average of RGB (depth or IR) features over time up to $k$-frame}. \textcolor{blue}{sergio: and $\delta$ is ...}

By optimizing Eq.~\ref{Eq:ranksvm}, we map a sequence of $K$ frames to a single vector ${\bf d}$. In this paper, rank pooling is directly applied on the pixels of RGB (Depth or IR) frames and the dynamic image ${\bf d}$ is of the same size as the input frames. In our case, given input frame, we compute its dynamic image online with rank pooling using $K$ consecutive frames. Our selection of dynamic images for rank pooling in SD-Net is further motivated by the fact that dynamic images have proved its superiority to regular optical flow~\cite{wang2017ordered,fernando2017rank}.

{\flushleft \textbf{Single-modal SD-Net.}}
%For any single-modal input frame, a corresponding dynamic image is online constructed with rank pooling using consecutive $N$ frames.
%which provides a complete face contour view and the motion clues of the current input in the next period of time. In fact, the static image has rich appearance details of the current state of the face, while the dynamic image owes more change information between different states.
As shown in Fig.~\ref{fig:static_dynamic}, taking the RGB modality as an example, we propose the SD-Net to learn hybrid features from static and dynamic images. It contains $3$ branches: static, dynamic and static-dynamic branches, which learn complementary features. The network takes ResNet-18~\cite{he2016deep} as the backbone. For static and dynamic branches, each of them consists of $5$ blocks (\ie, conv, res1, res2, res3, res4) and $1$ Global Average Pooling (GAP) layer, while in the static-dynamic branch, the conv and res1 blocks are removed because it takes fused features of res1 blocks from static and dynamic branches as input.

For convenience of terminology with the rest of the paper, we divide residual blocks of the network into a set of modules $ \{ {\cal M}^{t}_{\kappa} \}_{t=1}^{4}$ according to feature level, where $\kappa \in \{ color, depth, ir \}$ is an indicator of the modality and $t$ represents the feature level. Except for the first module ${\cal M}^{1}_{\kappa}$, each module extracts static, dynamic and static-dynamic features by using a residual block, denoted as ${\bf X}^{t}_{s,\kappa}$, ${\bf X}^{t}_{d,\kappa}$ and ${\bf X}^{t}_{f,\kappa}$, respectively.
The output features from each module are used as the input for the next module. The static-dynamic features ${\bf X}^{1}_{f,\kappa}$ of the first module are obtained by directly summing ${\bf X}^{1}_{s,\kappa}$ and ${\bf X}^{1}_{d,\kappa}$.

%For the first module, it only takes static and dynamic features after the convolutional and pooling layers as the inputs,and generate ${\bf X}^{1}_{s,\kappa}$, ${\bf X}^{1}_{d,\kappa}$ and ${\bf X}^{1}_{f,\kappa}$ three features. However, there is no residual block to extract ${\bf X}^{1}_{f,\kappa}$ specifically, where we obtain ${\bf X}^{1}_{f,\kappa}$ directly by using a sum of ${\bf X}^{1}_{s,\kappa}$ and ${\bf X}^{1}_{d,\kappa}$.

In order to ensure each branch learns independent features, each branch employs an independent loss function after the GAP layer~\cite{tan-ijcai}. In addition, a loss function based on the summed features from all three branches is employed. The binary cross-entropy loss is used as the loss function. All branches are jointly and concurrently optimized to capture discriminative and complementary features for face anti-spoofing in image sequences.
The overall objective function of SD-Net for the $\kappa^{th}$ modality is defined as:
\begin{equation}\label{Eq:single_modality_loss}
\begin{split}
\mathcal{L}^{\kappa}=  \mathcal{L}_{s}^{\kappa} + \mathcal{L}_{d}^{\kappa} + \mathcal{L}_{f}^{\kappa} + \mathcal{L}_{sdf}^{\kappa}
\end{split}
\end{equation}
where $\mathcal{L}_{s}^{\kappa}$, $\mathcal{L}_{d}^{\kappa}$, $\mathcal{L}_{f}^{\kappa}$ and $\mathcal{L}_{sdf}^{\kappa}$
are the losses for static branch, dynamic branch, static-dynamic branch, and summed features from all three branches of the network, respectively.

\begin{figure}[t]
\centering
\setlength{\tabcolsep}{8pt}
\includegraphics[width=1.0\linewidth]{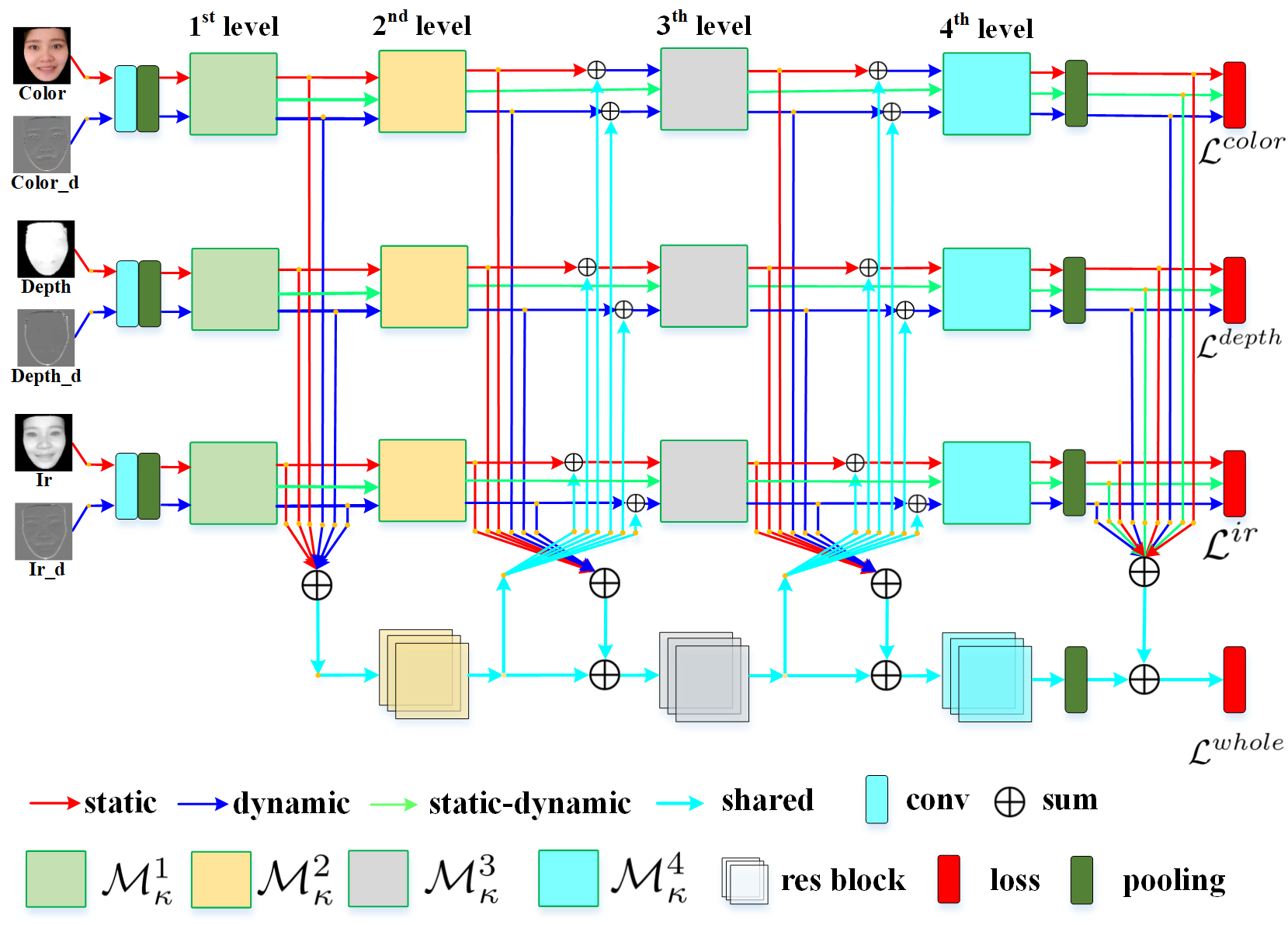}
\caption{The PSMM-Net diagram. It consists of two main parts. The first is the modality-specific network, which contains three SD-Nets to learn features from RGB, Depth, IR modalities, respectivel. The second is a shared branch for all modalities, which aims to learn the complementary features among different modalities.}
\label{fig:modal_fusion}
\end{figure}

\subsection{PSMM-Net for Multi-modal Fusion}
%To take full advantage of the multi-modal (\ie{RGB, Depth, IR}) information in CeFA dataset, such as RGB modality has rich appearance details, Depth modality measure the distance of each pixel from the face to camera, and IR modality response to the amount of face thermal radiation, prior works proposed fusion ways to fuse the complementary information between the three modalities.
\iffalse
To take full advantage of multi-modal (\ie{RGB, Depth, IR}) information in CASIA-SURF CeFA, such as RGB modality has rich appearance details, Depth modality measure the distance of each pixel from the face to camera, and IR modality response to the amount of face thermal radiation, we propose the PSMM-Net. %inspired by~\cite{Cao2018Partially}.
\fi
The architecture of the proposed PSMM-Net  is shown in Fig.~\ref{fig:modal_fusion}. It consists of two main parts: a) the modality-specific network, which
contains three SD-Nets to learn features from RGB, Depth, IR modalities, respectively; %with each one to learn the specific %semantics for one pair images of RGB, depth, IR and their corresponding dynamic images;
b) and a shared branch for all modalities, which aims to learn the complementary features among different modalities.

%For SD-Net, its detailed structure has been clarified above.
For the shared branch, we adopt ResNet-18, removing the first conv layer and res1 block. %(res2 block, res3 block, res4 block and GAP layer are retained).
In order to capture correlations and complementary semantics among different modalities, information exchange and interaction among SD-Nets and the shared branch are designed. This is done in two different ways: a) forward feeding of fused SD-Net features to the shared branch, and b) backward feeding from shared branch modules output to SD-Net block inputs.

\textbf{Forward Feeding}. We fuse static and dynamic SD-Nets features from all modality branches and fed them as input to its corresponding shared block.
The fused process at $t^{th}$ feature level can be formulated as:
\begin{equation}\label{Eq:sum1}
  {\tilde {\bf S}}^{t} = \sum_{\kappa }{ {\bf X}}^{t}_{s,\kappa}   + \sum_{\kappa }{ {\bf X}}^{t}_{d,\kappa} + {\bf S}^t \quad t=1,2,3
\end{equation}
In the shared branch, ${\tilde {\bf S}}^{t}$ denotes the input to the $(t + 1)^{th}$ block,
and ${\bf S}^t$ denotes the output of the $t^{th}$ block.
Note that the first residual block is removed from the shared branch, thus ${\bf S}^1$ equals to zero.

\textbf{Backward Feeding}. Shared features ${\bf S}^t$ are delivered back to the SD-Nets of the different modalities.
The static features ${ {\bf X}}^{t}_{s,\kappa}$ and dynamic features ${ {\bf X}}^{t}_{d,\kappa}$
add with ${\bf S}^t$ for feature fusion. This can be denoted as:
\begin{equation}\label{Eq:sum2}
  \widetilde{\bf{X}}_{s,\kappa}^t = {\bf{X}}_{s,\kappa}^t + {{\bf{S}}^t}, \quad \widetilde{\bf{X}}_{d,\kappa}^t = {\bf{X}}_{d,\kappa}^t + {{\bf{S}}^t}
\end{equation}
where $t$ ranges from 2 to 3.

After feature fusion, ${ \widetilde {\bf X}}^{t}_{s,\kappa}$ and ${ \widetilde {\bf X}}^{t}_{d,\kappa}$ become the new static and dynamic features, which are then feed to the next module ${\cal M}^{t+1}_{\kappa}$. Note that the exchange and interaction among SD-Nets and the shared branch are only performed for static and dynamic features. This is done to avoid hybrid features among static and dynamic information to be disturbed by multi-modal semantics.

\textbf{Loss Optimization}. There are two main kind of losses employed to guide the training of PSMM-Net. The first corresponds to the losses of the three SD-Nets, \ie. color, depth and ir modalities, denoted as ${\cal L}^{color}$, ${\cal L}^{depth}$ and ${\cal L}^{ir}$, respectively.
The second corresponds to the loss that guides the entire network training, denoted as ${\cal L}^{whole}$,
 which bases on the summed features from all SD-Nets and the shared branch. The overall loss $\cal L$ of PSMM-Net is denoted as:
\begin{equation}\label{Eq:multi_modality_loss}
\begin{split}
\mathcal{L}= \mathcal{L}^{whole} + \mathcal{L}^{color} +  \mathcal{L}^{depth} + \mathcal{L}^{ir}
\end{split}
\end{equation}

\section{CASIA-SURF CeFA dataset}

This section describes the CASIA-SURF CeFA dataset. The motivation of this dataset is to provide with an increased diversity of attack types compared to existing datasets, as well as to explore the effect of cross-ethnicity in face anti-spoofing, which has received little attention in the literature. Furthermore, it contains three visual modalities, \ie., RGB, Depth, and IR. Summarizing, the main purpose of CASIA-SURF CeFA is to provide with the largest up to date face anti-spoofing dataset to allow for the evaluation of the generalization performance of new PAD methods in three main aspects: cross-ethnicity, cross-modality and cross-attacks. In this section, we describe the CASIA-SURF CeFA dataset in detail, including acquisition details, attack types, and proposed evaluation protocols.

\begin{figure}[htb]
	\begin{center}
		\includegraphics[width=0.8\linewidth]{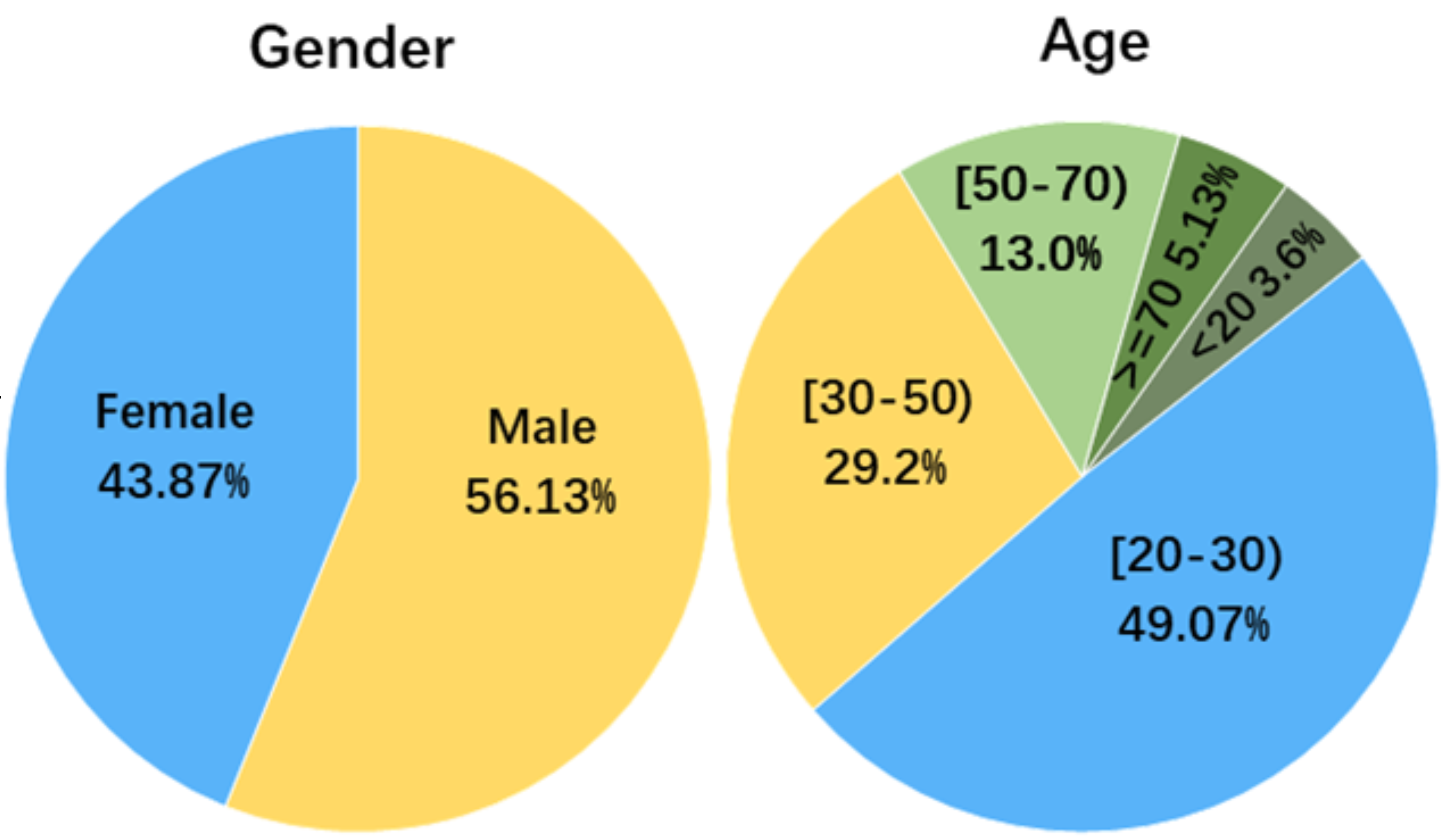}
	\end{center}
	\caption{Age and gender distributions of the CASIA-SURF CeFA.}
	\label{gender_age}
	% \label{fig:onecol}
\end{figure}

\begin{table*}[]
\scalebox{0.82}{
	\begin{tabular}{|c|c|c|c|c|c|c|c|c|c|c|c|c|c|}
		\hline
		Prot. & Subset & \multicolumn{3}{c|}{Ethnicity} & Subjects & \multicolumn{3}{c|}{Modalities} & \multicolumn{2}{c|}{PAIs} & \# real videos & \# fake videos & \# all videos \\ \hline \hline
		\multicolumn{2}{|c|}{} & 1\_1 & 1\_2 & 1\_3 & \multicolumn{9}{c|}{} \\ \hline
		\multirow{3}{*}{1} & Train & A & C & E & 1-200 & \multicolumn{3}{c|}{R\&D\&I} & \multicolumn{2}{c|}{Pr\&Re} & 600/600/600 & 1800/1800/1800 & 2400/2400/2400 \\ \cline{2-14}
		& Valid & A & C & E & 201-300 & \multicolumn{3}{c|}{R\&D\&I} & \multicolumn{2}{c|}{Pr\&Re} & 300/300/300 & 900/900/900 & 1200/1200/1200 \\ \cline{2-14}
		& Test & C\&E & A\&E & A\&C & 301-500 & \multicolumn{3}{c|}{R\&D\&I} & \multicolumn{2}{c|}{Pr\&Re} & 1200/1200/1200 & 6600/6600/6600 & 7800/7800/7800 \\ \hline
		\multicolumn{9}{|c|}{} & 2\_1 & 2\_2 & \multicolumn{3}{c|}{} \\ \hline
		\multirow{3}{*}{2} & Train & \multicolumn{3}{c|}{A\&C\&E} & 1-200 & \multicolumn{3}{c|}{R\&D\&I} & Pr & Re & 1800/1800 & 3600/1800 & 5400/3600 \\ \cline{2-14}
		& Valid & \multicolumn{3}{c|}{A\&C\&E} & 201-300 & \multicolumn{3}{c|}{R\&D\&I} & Pr & Re & 900/900 & 1800/900 & 2700/1800 \\ \cline{2-14}
		& Test & \multicolumn{3}{c|}{A\&C\&E} & 301-500 & \multicolumn{3}{c|}{R\&D\&I} & Pe & Pr & 1800/1800 & 4800/6600 & 6600/8400 \\ \hline
		\multicolumn{6}{|c|}{} & 3\_1 & 3\_2 & 3\_3 & \multicolumn{5}{c|}{} \\ \hline
		\multirow{3}{*}{3} & Train & \multicolumn{3}{c|}{A\&C\&E} & 1-200 & R & D & I & \multicolumn{2}{c|}{Pr\&Re} & 600/600/600 & 1800/1800/1800 & 2400/2400/2400 \\ \cline{2-14}
		& Valid & \multicolumn{3}{c|}{A\&C\&E} & 201-300 & R & D & I & \multicolumn{2}{c|}{Pr\&Re} & 300/300/300 & 900/900/900 & 1200/1200/1200 \\ \cline{2-14}
		& Test & \multicolumn{3}{c|}{A\&C\&E} & 301-500 & D\&I & R\&I & R\&D & \multicolumn{2}{c|}{Pr\&Re} & 1200/1200/1200 & 5600/5600/5600 & 6800/6800/6800 \\ \hline
		\multicolumn{2}{|c|}{} & 4\_1 & 4\_2 & 4\_3 & \multicolumn{9}{c|}{} \\ \hline
		\multirow{3}{*}{4} & Train & A & C & E & 1-200 & R & D & I & \multicolumn{2}{c|}{Re} & 600/600/600 & 600/600/600 & 1200/1200/1200 \\ \cline{2-14}
		& Valid & A & C & E & 201-300 & R & D & I & \multicolumn{2}{c|}{Re} & 300/300/300 & 300/300/300 & 600/600/600 \\ \cline{2-14}
		& Test & C\&E & A\&E & A\&C & 301-500 & R & D & I & \multicolumn{2}{c|}{Pr} & 1200/1200/1200 & 5400/5400/5400 & 6600/6600/6600 \\ \hline
	\end{tabular}
	}
\caption{Four evaluation protocols are defined for CASIA-SURF CeFA; 1) cross-ethnicity, 2) cross-PAI, 3) cross-modality and 4) cross-ethnicity\&PAI, respectively. %$500$ subjects per ethnicity are divided into train, valid and testing sets.
	Note that 3D attacks subset of CASIA-SURF CeFA are included to the test set of every testing protocol (not shown in the table). R: RGB, D: Depth, I: IR, A: Africa, C: Central Asia, E: East Asia, Pr: print attack, Re: replay attack; \& indicates merging; $*\_*$ corresponds to the name of sub-protocols.}
\label{tab:protocol}
\end{table*}

\textbf{Acquisition Details.} We use the Intel Realsense to capture the RGB, Depth and IR videos simultaneously at 30 fps. The resolution is 1280 $\times$ 720 pixels for each video frame and all modalities. Performers are asked to move smoothly their head so as to have a maximum of around $30^0$ deviation of head pose in relation to frontal view. Data pre-processing is similar to the one performed in~\cite{DBLP:conf/cvpr/abs-1812-00408}, expect that PRNet~\cite{DBLP:conf/eccv/FengWSWZ18} is replaced by 3DFFA~\cite{zhu2017face} for face region detection. Examples of original recorded images from video sequences and processed face regions for different visual modalities are shown in Fig.~\ref{samp_data}.

\textbf{Statistics.} As shown in Table~\ref{tab1}, CASIA-SURF CeFA consists of 2D and 3D attack subsets. For the 2D attack subset, it includes print and video-reply attacks, and three ethnicites (African, East Asian and Central Asian) with 2 attacks (print face from cloth and video-replay). Each ethnicity has 500 subjects. Each subject has 1 real sample, 2 fake samples of print attack captured in indoor and outdoor, and 1 fake sample of video-replay. In total, there are 18000 videos (6000 per modality). The age and gender statistics for the 2D attack subset of CASIA-SURF CeFA is shown in Fig.~\ref{gender_age}. %It shows that the ratio of male about $56.13\%$ and most of peoples are between 20 and 50 years old.

For the 3D attack subset, it has 3D print mask and silica gel face attacks. For 3D print mask, it has 99 subjects, each subject with 18 fake samples captured in three attacks and six lighting environments. Attacks include only mask, wearing a wig and glasses, and wearing a wig and no glasses. Lighting conditions include outdoor sunshine, outdoor shade, indoor side light, indoor front light, indoor backlit and indoor regular light. In total, there are 5346 videos (1782 per modality). For silica gel face attacks, it has 8 subjects, each subject has 8 fake samples captured in two attacks styles and four lighting environments. Attacks include wearing a wig and glasses and wearing a wig and no glasses. Lighting environments include indoor side light, indoor front light, indoor backlit and indoor normal light. In total, there are 196 videos (64 per modality).

\textbf{Evaluation Protocols.}
We design four protocols for the 2D attacks subset, as shown in Table~\ref{tab:protocol}, totalling 11 sub-protocols (1\_1, 1\_2, 1\_3, 2\_1, 2\_2, 3\_1, 3\_2, 3\_3, 4\_1, 4\_2, and 4\_3).  We divide $500$ subjects per ethnicity into three subject-disjoint subsets (second and fourth columns in Table~\ref{tab:protocol}). Each protocol has three data subsets: training, validation and testing sets, which contain 200, 100, and 200 subjects, respectively.
%The training set has 200 subjects per ethnicity to train the model. The validation set has 100 subjects, to be used for model selection. The testing set is used for final evaluation, including 200 subjects. The 3D attack subset of CeFA is included in the test set.\\\\
\iffalse
Given that the data size of the 3D attacks subset (~107 subjects) is smaller than the 2D attacks subset (~1500 subjects) and the unseen attack is more challenging and reasonable for real applications. Therefore, we add the 3D attack subset as one part of the testing set in four protocols.
 \fi
\\
\textbf{$\bullet$ Protocol 1 (cross-ethnicity)}: Most of the public face PAD datasets just contain a single ethnicity. Even though there are few datasets~\cite{Liu2018Learning,DBLP:conf/cvpr/abs-1812-00408} containing multiple ethnicities, they lack of ethnicity labels or do not provide with a protocol to perform cross-ethnicity evaluation. Therefore, we design the first protocol to evaluate the generalization of PAD methods for cross-ethnicity testing.
One ethnicity is used for training and validation, and the left two ethnicities are used for testing. Therefore, there are three different evaluations (third column of Protocol 1 in Table~\ref{tab:protocol}.\\
\textbf{$\bullet$ Protocol 2 (cross-PAI)}: Given the diversity and unpredictability of attack types from different presentation attack instruments (PAI), it is necessary to evaluate the robustness of face PAD algorithms to this kind of variations (sixth column of Protocol 2 in Table.~\ref{tab:protocol}). \\
\textbf{$\bullet$ Protocol 3 (cross-modality)}: Given the release of affordable devices capturing complementary visual modalities (\ie, Intel Resense, Mircrosoft Kinect), recently the multi-modal face anti-spoofing dataset was proposed~\cite{DBLP:conf/cvpr/abs-1812-00408}. However, there is no standard protocol to explore the generalization of face PAD methods when different train-test modalities are considered for evaluation. We define three cross-modality  evaluations, each of them having one modality for training and the two remaining ones for testing (fifth column of Protocol 3 in Table.~\ref{tab:protocol}). \\
\textbf{$\bullet$ Protocol 4 (cross-ethnicity \& PAI)}: The most challenging protocol is designed via combining the condition of both Protocol 1 and 2. As shown in Protocol 4 of Table.~\ref{tab:protocol}, the testing subset introduces two unknown target variations simultaneously.

\iffalse
Thus, we define this protocol to fill the gap out of two feasibility analysis: (1) there is a certain mapping relationship between multi-modalities for the same category of samples. Such as for real sample, the color modality owes realistic texture details, the Depth modality characterize different texture changes according to the line of the face organ, and the IR modality depicts the thermal imaging according to the amount of heat radiated from a face. These do not exist in 2D presentation attacks. (2) Deep neural networks have the ability to learn this mapping and promote multi-modal information complementary.
\fi

\iffalse
In order to increase the difficulty to cope with the complex authentication scenarios, we use the ``Leave-One-Variation-Out'' in all sub-protocols as the method evaluation.
\fi
Like ~\cite{Boulkenafet2017OULU}, the mean and variance of evaluate metrics for these four protocols are calculated in our experiments. Detailed statistics for the different protocols are shown in Table~\ref{tab:protocol}. More information about CASIA-SURF CeFA can be found in our supplementary material.

\begin{figure}[t]
	\centering
	\setlength{\tabcolsep}{10pt}
	\includegraphics[width=1.0\linewidth]{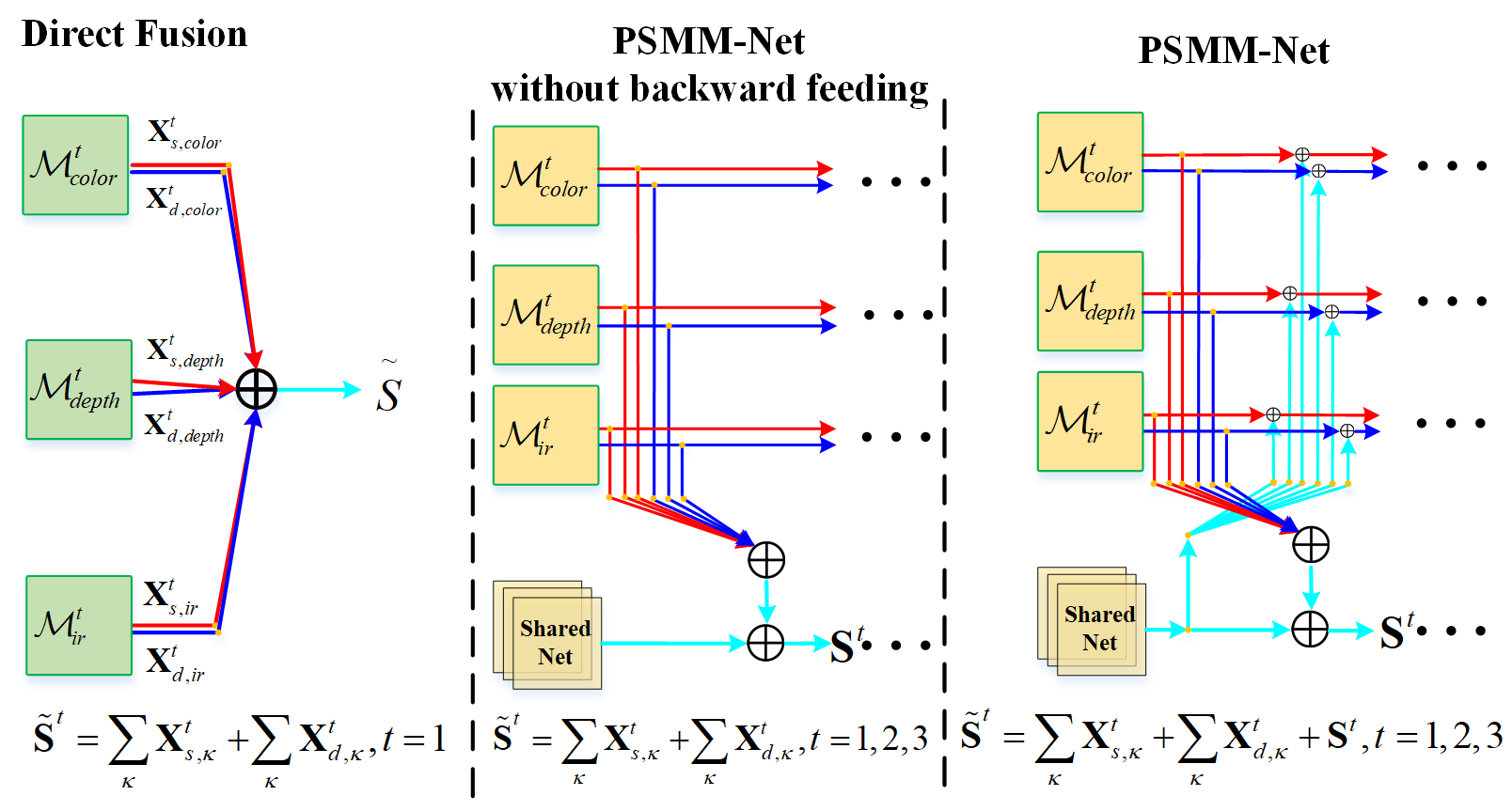}
	\caption{Comparison of network units for multi-modal fusion strategies. From left to right: NHF, PSMM-NET-WoBF and PSMM-Net. The fusion process for the $t^{th}$ feature level of each strategy is shown at the bottom.}
	\label{fig:fusion_ways}
\end{figure}

\section{Experiments}
In this section, we conduct a series of experiments on public available face anti-spoofing datasets to verify the effectiveness of our methodology and the benefits of the presented CASIA-SURF CeFA dataset. In the following, we will introduce the employed datasets \& metrics, implementation details, experimental setting, and results \& analysis sequentially.

\subsection{Datasets \& Metrics}
We evaluate the performance of PSMM-Net on two multi-modal (\ie, RGB, Depth and IR) datasets: CASIA-SURF CeFA and CASIA-SURF~\cite{DBLP:conf/cvpr/abs-1812-00408}, while evaluate the SD-Net on two single-modal (\ie, RGB) face anti-spoofing benchmarks: OULU-NPU~\cite{Boulkenafet2017OULU} and SiW~\cite{Liu2018Learning}. They are the mainstream datasets released in recent years with their own characteristics in terms of the number of subject, modality and ethnicity, attack types, acquisition device and PAIs~\etal. Therefore, experiments on these datasets can verify the performance of our method more convincingly.

In order to perform a consistent evaluation with prior works, we report the experimental results using the following metrics based on respective official protocols: Attack Presentation Classification Error Rate (APCER)~\cite{metrics}, Bona Fide Presentation Classification Error Rate (BPCER), Average Classification Error Rate (ACER), and Receiver Operating Characteristic (ROC) curve~\cite{DBLP:conf/cvpr/abs-1812-00408}.

\subsection{Implementation Details}
The proposed PSMM-Net is implemented with Tensorflow~\cite{AbadiTensorFlow} and run on a single NVIDIA TITAN X GPU. We resize the cropped face region to $112\times112$, and use random rotation within the range of [$-180^0$, $180^0$], flipping, cropping and color distortion for data augmentation. All models are trained for $25$ epochs via Adaptive Moment Estimation (Adam) algorithm and initial learning rate of $0.1$, which is decreased after $15$ and $20$ epochs with a factor of $10$. The batch size of each CNN stream is $64$, and the length of the consecutive frames used to construct dynamic map is set to $7$ by our experimental experience. In addition, all fusion points in this work use element summation operations to prevent dimension explosion.

\begin{table}[]\centering
\setlength\tabcolsep{4pt}
%\tiny
%\scriptsize
%\footnotesize
\small
\begin{tabular}{|c|c|c|c|c|}
\hline
\multicolumn{2}{|c|}{Prot. name}        & APCER(\%)    & BPCER(\%)   & ACER(\%)     \\ \hline \hline
\multirow{4}{*}{Prot. 1} & 1\_1     & 0.5      & 0.8     & 0.6      \\ \cline{2-5}
                            & 1\_2     & 4.8      & 4.0     & 4.4      \\ \cline{2-5}
                            & 1\_3     & 1.2      & 1.8     & 1.5      \\ \cline{2-5}
                            & Avg$\pm$Std & 2.2$\pm$2.3  & 2.2$\pm$1.6 & 2.2$\pm$2.0  \\ \hline
\multirow{3}{*}{Prot. 2} & 2\_1     & 0.1      & 0.7     & 0.4      \\ \cline{2-5}
                            & 2\_2     & 13.8     & 1.2     & 7.5      \\ \cline{2-5}
                            & Avg$\pm$Std & 7.0$\pm$9.7  & 1.0$\pm$0.4 & 4.0$\pm$5.0  \\ \hline
\multirow{4}{*}{Prot. 3} & 3\_1     & 8.9      & 0.9     & 4.9      \\ \cline{2-5}
                            & 3\_2     & 22.6     & 4.6     & 13.6     \\ \cline{2-5}
                            & 3\_3     & 21.1     & 2.3     & 11.7     \\ \cline{2-5}
                            & Avg$\pm$Std & 17.5$\pm$7.5 & 2.6$\pm$1.9 & 10.1$\pm$4.6 \\ \hline
\multirow{4}{*}{Prot. 4} & 4\_1    & 33.3      & 15.8     & 24.5      \\ \cline{2-5}
                            & 4\_2     & 78.2      & 8.3     & 43.2      \\ \cline{2-5}
                            & 4\_3     & 50.0      & 5.5     & 27.7      \\ \cline{2-5}
                            & Avg$\pm$Std & 53.8$\pm$22.7  & 9.9$\pm$5.3 & 31.8$\pm$10.0  \\ \hline
\end{tabular}
\caption{PSMM-Net evaluation on the four protocols of CASIA-SURF CeFA dataset, where A$\_$B represents sub-protocol B from Protocol A, and Avg$\pm$Std indicates the mean and variance operation.}
\label{tab:sto_race}
\end{table}

\begin{table*}[]\setlength\tabcolsep{4pt}
	%\tiny
	%\scriptsize
	%\footnotesize
	\small
	\centering
	\begin{tabular}{|c|c|c|c|c|c|c|c|c|c|}
		\hline
		\multirow{2}{*}{Prot.1}
		& \multicolumn{3}{c|}{RGB}
		& \multicolumn{3}{c|}{Depth}
		& \multicolumn{3}{c|}{IR} \\ \cline{2-10}
		& APCER($\%$)  & BPCER($\%$)  & ACER($\%$)  & APCER($\%$)   & BPCER($\%$)
		& ACER($\%$)   & APCER($\%$)  & BPCER($\%$)  & ACER($\%$)  \\ \hline \hline
		S-Net & 28.1$\pm$3.6          & \textbf{6.4$\pm$4.6}  & 17.2$\pm$3.6
		& \textbf{5.6$\pm$3.0}  & 9.8$\pm$4.2           & 7.7$\pm$3.5
		& 11.4$\pm$2.1          & 8.2$\pm$1.2           & 9.8$\pm$1.7  \\ \hline
		
		D-Net  & 20.6$\pm$4.0         & 19.3$\pm$9.0          & 19.9$\pm$4.0
		& 11.2$\pm$5.1         & 7.5$\pm$1.5           & 9.4$\pm$2.0
		& 8.1$\pm$1.8          & 14.4$\pm$3.8          & 11.3$\pm$2.1  \\ \hline
		
		SD-Net  & \textbf{14.9$\pm$6.0}         & 10.3$\pm$1.8                  & \textbf{12.6$\pm$3.4}
		& 7.0$\pm$8.1                   & \textbf{5.2$\pm$3.5}          & \textbf{6.1$\pm$5.4}
		& \textbf{7.3$\pm$1.2}          & \textbf{5.5$\pm$1.8}          & \textbf{6.4$\pm$1.3}  \\ \hline
	\end{tabular}
	\caption{Ablation experiments on three single-modal groups: RGB, Depth and IR. Each modality group contains three experiments: static branch, dynamic branch and static-dynamic branch. %For all performance metrics (\ie, APCER, BPCER or ACER), the
		Numbers in bold correspond to the best results per column.}
	\label{tab:casia_race_ablation_SDNet}
\end{table*}

%\begin{figure}[t]
%\centering
%\setlength{\tabcolsep}{10pt}
%\includegraphics[width=1.0\linewidth]{D:/CVPR2020/NEW CASIA-SURF2.0/latex/image/fusion_ways.png}
%\caption{Comparison of network units for multi-modal fusion strategies. From left to right: NHF, PSMM-NET-WoBF and PSMM-Net. The fusion process for the $t^{th}$ feature level of each strategy is shown at the bottom.}
%\label{fig:fusion_ways}
%\end{figure}

\begin{table*}[]
	%\tiny
	%\scriptsize
	\footnotesize
	%\small
	\centering
	\begin{tabular}{|c|c|c|c|c|c|c|c|c|c|}
		\hline
		\multirow{2}{*}{Prot. name} & \multicolumn{3}{c|}{RGB}  & \multicolumn{3}{c|}{Depth} & \multicolumn{3}{c|}{IR}          \\ \cline{2-10}
		& APCER(\%) & BPCER(\%) & (ACER\%)  & APCER(\%) & BPCER(\%) & (ACER\%) & APCER(\%) & BPCER(\%) & (ACER\%) \\ \hline
		\hline
		Prot. 1  & 14.9$\pm$6.0  & 10.3$\pm$1.8  & 12.6$\pm$3.4  & 7.0$\pm$8.1   & 5.2$\pm$3.5
		& 6.1$\pm$5.4  & 7.3$\pm$1.2   & 5.5$\pm$1.8   & 6.4$\pm$1.3
		\\ \hline
		Prot. 2  & 45.0$\pm$39.1 & 1.6$\pm$1.9   & 23.3$\pm$18.6 & 13.6$\pm$18.7 & 1.2$\pm$0.7
		& 7.4$\pm$9.7  & 8.1$\pm$11.0  & 1.5$\pm$1.8   & 4.8$\pm$6.4
		\\ \hline
		Prot. 3$^{*}$  & 5.9       & 2.2       & 4.0       & 0.3       & 0.3
		& 0.3      & 0.2       & 0.5       & 0.4
		\\ \hline
		Prot. 4  & 65.8$\pm$16.4 & 8.3$\pm$6.5   & 35.2$\pm$5.8  & 18.5$\pm$8.2  & 7.0$\pm$5.2
		& 12.7$\pm$5.7 & 6.8$\pm$2.9   & 4.2$\pm$3.3   & 5.5$\pm$2.2
		\\ \hline
	\end{tabular}
	\caption{Experimental results of the SD-Net based on single modality on four protocols ($*$ indicates that the modal type of the testing subset is consistent with the training subset).}
	\label{tab:single_modal_results}
\end{table*}

\begin{table}[]
\scalebox{0.90}{
	\begin{tabular}{|c|c|c|c|}
		\hline
		\multirow{2}{*}{Prot.1} & \multicolumn{3}{c|}{PSMM-Net} \\ \cline{2-4}
		& APCER($\%$) & BPCER($\%$) & ACER($\%$) \\ \hline \hline
		RGB& 14.9$\pm$6.0 &10.3$\pm$1.8  & 12.6$\pm$3.4 \\ \hline
		RGB\&Depth &2.3$\pm$2.9  &9.2$\pm$5.9  & 5.7$\pm$3.5 \\ \hline
		RGB\&Depth\&IR &\textbf{2.2$\pm$2.3 }  &\textbf{2.2$\pm$1.6} &\textbf{2.2$\pm$2.0}  \\ \hline
	\end{tabular}
	}
\caption{ Ablation experiments on the effect of multiple modalities. %We evaluate one modality (RGB), two modalities (RGB and depth), and three modalities (RGB, depth and IR) on PSMM-Net for comparison.
Numbers in bold correspond to the best result per column.
}
\label{tab:casia_race_ablation}
\end{table}

\begin{table}[]
\scalebox{0.96}{
\begin{tabular}{|c|c|c|c|}
\hline
Method   & APCER(\%) & BPCER(\%) & ACER(\%) \\ \hline \hline
NHF      & 25.3$\pm$12.2  & 4.4$\pm$3.1   & 14.8$\pm$6.8  \\ \hline
PSMM-WoBF & 12.7$\pm$0.4  & 3.2$\pm$2.3   & 7.9$\pm$1.3  \\ \hline
PSMM-Net &  \textbf{2.2$\pm$2.3}  & \textbf{2.2$\pm$1.6}   & \textbf{2.2$\pm$2.0}  \\ \hline
\end{tabular}
}
\caption{Comparison of fusion strategies in Protocol 1 of CASIA-SURF CeFA. The number in black indicates best results.
%The numbers in bold are best results.
}
\label{tab:NHF_VS_pSMM-WBF_VS_pSMM}
\end{table}

\begin{table*}
	%\tiny
	%\scriptsize
	\footnotesize
	%\small
	\centering
	\setlength{\tabcolsep}{11pt}
	\scalebox{0.96}{
		\begin{tabular}{|c|c|c|c|c|c|c|}
			\hline
			\multirow{2}{*}{Method} & \multicolumn{3}{c|}{TPR (\%)} & \multirow{2}{*}{APCER (\%)} & \multirow{2}{*}{BPCER (\%)} & \multirow{2}{*}{ACER (\%)}\\
			\cline{2-4}
			& @FPR=$10^{-2}$ &@FPR=$10^{-3}$ &@FPR=$10^{-4}$ & &  & \\
			\hline
			\hline
			NHF fusion~\cite{DBLP:conf/cvpr/abs-1812-00408} &89.1 &33.6 &17.8 &5.6 &3.8 &4.7 \\ \hline
			Single-scale SE fusion~\cite{DBLP:conf/cvpr/abs-1812-00408} &96.7 &81.8 &56.8 &3.8 &1.0 &2.4\\ \hline
			Multi-scale SE fusion~\cite{zhang2019casiasurf} &99.8 &98.4 &95.2 &1.6 &0.08 &0.8 \\	\hline
			% VisionLabs & 99.9 & 99.9 & 99.8 & 0.0 & 0.1 & 0.0 \\ \hline
			% ReadSense & 100.0 & 99.9 & 99.8 & 0.1 & 0.0 & 0.0 \\ \hline
			% Feather & 99.9 & 99.8 & 98.1 & 0.1 & 0.1 & 0.1 \\ \hline
			% Feather & 99.9 & 99.8 & 98.1 & 0.1 & 0.1 & 0.1 \\ \hline
			% Hahahaha & 99.6 & 98.5 & 93.1 & 0.1 & 1.2 & 0.6 \\ \hline
			% MAC-adv-group & 99.5 & 97.2 & 89.5 & 2.0 & 0.1 & 1.1 \\ \hline
			% ZKBH & 99.7 & 96.8 & 87.6 & 0.9 & 0.2 & 0.5 \\ \hline
			% VisionMiracle & 99.9 & 98.3 & 87.2 & 0.2 & 0.4 & 0.3 \\ \hline
			% GrandiantResearch & 97.0 & 77.4 & 63.5 & 1.9 & 1.4 & 1.6 \\ \hline
			PSMM-Net       &\textbf{99.9} &99.3 &96.2 &0.7 &0.06 &0.4  \\	\hline
			PSMM-Net(CASIA-SURF CeFA) &\textbf{99.9} &\textbf{99.7} &\textbf{97.6} &\textbf{0.5} &\textbf{0.02} &\textbf{0.2}  \\	\hline
		\end{tabular}
	}
	\caption{Comparison of the proposed method with three fusion strategies. All models are trained on the CASIA-SURF training subset and tested on the testing subset. Best results are bolded.}
	\label{tab:casia_surf_result}
\end{table*}

\subsection{Baseline Model Evaluation}
Before exploring the traits of our dataset, we first provide a benchmark for CASIA-SURF CeFA based on the proposed method. From the Table~\ref{tab:sto_race}, in which the results of the four protocols are derived from all the respective sub-protocols by calculating the mean and variance, we can draw the following conclusions: (1) from the results of the three sub-protocols in Protocol 1, the ACER scores are $0.6\%$, $4.4\%$ and $1.5\%$, respectively, indicating that it is necessary to study the generalization of the face PAD method for different ethnicity; (2) In the case of Protocol 2, when print attack is used for training/validation and video-replay and 3D mask are used for testing, the ACER score is $0.4\%$ (sub-protocol 2\_1), while video-replay attack is used for training/validation, and print attack and 3D attack are used for testing, with an ACER score of $7.5\%$ (sub-protocol 2\_2). The large gap between the results of the two sub-protocols is mainly caused by different PAI (\ie. different displays and printers) create different artifacts. (3) Protocol 3 evaluates cross-modality. The best result is achieved for sub-protocol 3\_1, with ACER of $4.9\%$. The other two sub-protocols achieve a similar low performance score. This means the best performance is achieved when RGB data of 2D attack subset is used for training/validation while the other two modalities of 2D and 3D attack subsets are used for testing. (4) Protocol 4 is the most difficult evaluation scenario, which simultaneously considers cross-ethnicity and cross-PAI. All sub-protocols achieve poor performance, being $24.5\%$, $43.2\%$, and $27.7\%$ ACER scores for 4\_1, 4\_2, and 4\_3 achieve, respectively.

\subsection{Ablation Analysis}
In order to verify the effectiveness of the proposed method, we perform a series of ablation experiments on Protocol 1 (cross-ethnicity) of the CASIA-SURF CeFA dataset.

\textbf{Static and Dynamic Features.} We evaluate S-Net (Static branch of SD-Net), D-Net (Dynamic branch of SD-Net) and SD-Net. Results for RGB, Depth and IR modalities are shown in Table~\ref{tab:casia_race_ablation_SDNet}. Compared with S-Net and D-Net, SD-Net achieves superior performance. For RGB, Depth and IR modalities, ACER of SD-Net is $12.6\%$, $6.1\%$, $6.4\%$, versus $17.2\%$, $7.7\%$, $9.4\%$ of S-Net (improved by $4.6\%$, $1.6\%$, $3.4\%$) and $19.9\%$, $9.4\%$, $11.3\%$ of D-Net (improved by $7.3\%$, $3.3\%$, $4.9\%$), respectively. Furthermore, Table~\ref{tab:casia_race_ablation_SDNet} shows that the performance of Depth and IR modalities are superior to the one of RGB. One reason is the variability in lighting conditions included in CASIA-SURF CeFA. %Therefore, Table~\ref{tab:casia_race_ablation_SDNet} has demonstrated not only the effectiveness of SD-Net but also the advantages of depth and IR modalities for face anti-spoofing.

In addition, we provide the results of single-modal experiments on the $4$ protocols to facilitate comparison of face PAD algorithms, shown in Table~\ref{tab:single_modal_results}. It shows that when only single modality is used, the performance of the depth or IR modality is superior to that of the RGB modality.

\textbf{Multiple Modalities.}  In order to show the effect of analysing a different number of modalities, we evaluate one modality (RGB), two modalities (RGB and Depth), and three modalities (RGB, Depth and IR) on PSMM-Net. As shown in Fig.~\ref{fig:modal_fusion}, the PSMM-Net contains three SD-Nets and one shared branch. When only RGB modality is considered, we just use one SD-Net for evaluation. When two or three modalities are considered, we use two or three SD-Nets and one shared branch to train the PSMM-Net model, respectively. Results are shown in Table~\ref{tab:casia_race_ablation}. The best results are obtained when using all three modalities, which $2.2\%$ of APCER, $2.2\%$ of BPCER and $2.2\%$ of ACER. Compared with the performance of using single RGB modality and two modalities, the improvement in performance corresponds to $12.7\%$ and $0.1\%$ for APCER, $8.1\%$ and $7.2\%$ for BPCER, and $10.4\%$ and $3.5\%$ for ACER, respectively.
%\textbf{Rank pooling vs. optical flow.} balabalba....

% \begin{table}[]
% \scalebox{0.90}{
% 	\begin{tabular}{|c|c|c|c|}
% 		\hline
% 		\multirow{2}{*}{Prot.1} & \multicolumn{3}{c|}{PSMM-Net} \\ \cline{2-4}
% 		& APCER($\%$) & BPCER($\%$) & ACER($\%$) \\ \hline \hline
% 		RGB& 14.9$\pm$6.0 &10.3$\pm$1.8  & 12.6$\pm$3.4 \\ \hline
% 		RGB\&Depth &2.3$\pm$2.9  &9.2$\pm$5.9  & 5.7$\pm$3.5 \\ \hline
% 		RGB\&Depth\&IR &\textbf{2.2$\pm$2.3 }  &\textbf{2.2$\pm$1.6} &\textbf{2.2$\pm$2.0}  \\ \hline
% 	\end{tabular}
% 	}
% \caption{ Ablation experiments on the effect of multiple modalities. %We evaluate one modality (RGB), two modalities (RGB and depth), and three modalities (RGB, depth and IR) on PSMM-Net for comparison.
% Numbers in bold correspond to the best result per column.
% }
% \label{tab:casia_race_ablation}
% \end{table}

% \begin{table}[]
% \scalebox{0.96}{
% \begin{tabular}{|c|c|c|c|}
% \hline
% Method   & APCER(\%) & BPCER(\%) & ACER(\%) \\ \hline \hline
% NHF      & 25.3$\pm$12.2  & 4.4$\pm$3.1   & 14.8$\pm$6.8  \\ \hline
% PSMM-WoBF & 12.7$\pm$0.4  & 3.2$\pm$2.3   & 7.9$\pm$1.3  \\ \hline
% PSMM-Net &  \textbf{2.2$\pm$2.3}  & \textbf{2.2$\pm$1.6}   & \textbf{2.2$\pm$2.0}  \\ \hline
% \end{tabular}
% }
% \caption{Comparison of fusion strategies in Protocol 1 of CASIA-CeFA. The number in black indicates best results.
% %The numbers in bold are best results.
% }
% \label{tab:NHF_VS_pSMM-WBF_VS_pSMM}
% \end{table}

\textbf{Fusion Strategy.}
In order to evaluate the performance of PSMM-Net, we compare it with other two variants:  Naive halfway fusion (NHF) and PSMM-Net without backward feeding mechanism (PSMM-Net-WoBF). As shown in Fig.~\ref{fig:fusion_ways}, NHF combines the modules of different modalities at a later stage (\ie, after ${\cal M}^{1}_{\kappa}$ module) and PSMM-Net-WoBF strategy removes the backward feeding from PSMM-Net. The fusion comparison results are shown in Table~\ref{tab:NHF_VS_pSMM-WBF_VS_pSMM}, showing higher performance of the proposed PSMM-Net, with ACER of $2.2\%$.

\subsection{Methods Comparison}
\textbf{CASIA-SURF Dataset.}  Comparison results are show in Table~\ref{tab:casia_surf_result}. The performance of the PSMM-Net is superior to the ones of the competing multi-modal fusion methods, including Halfway fusion~\cite{DBLP:conf/cvpr/abs-1812-00408}, single-scale SE fusion~\cite{DBLP:conf/cvpr/abs-1812-00408}, and multi-scale SE fusion~\cite{zhang2019casiasurf}. When compared with~\cite{DBLP:conf/cvpr/abs-1812-00408,zhang2019casiasurf}, PSMM-Net improves the performance by at least $0.9\%$ for APCER, $0.02\%$ for NPECE, and $0.4\%$ for ACER. When the PSMM-Net is pretrained on CASIA-SURF CeFA, it further improves performance. Concretely, the performance of $TPR@FPR=10^{-4}$ is increased by $2.4\%$ when pretraining with the proposed CASIA-SURF CeFA dataset. %The improvement indicates that pre-training on the proposed dataset supports the generalization.

In 2019 a challenge on the CASIA-SURF dataset was run at CVPR~\footnote{\url{https://sites.google.com/qq.com/face-anti-spoofing/welcome/challengecvpr2019?authuser=0}}. The results of the challenge were very promising, where $3$ winning teams VisionLab~\cite{parkin2019recognizing}, ReadSense~\cite{shen2019facebagnet} and Feather~\cite{zhang2019feathernets} got TPR=$99.87\%@FPR=10^{-4}$, $99.81\%@FPR=10^{-4}$ and $99.14\%@FPR=10^{-4}$, respectively. The $2$ main reasons of these high performance are: 1) \emph{several external datasets were used.} VisionLab~\cite{parkin2019recognizing} used four lare-scale datasets, namely CASIA-WebFace~\cite{yi2014learning}, MSCeleb-1M~\cite{guo2016ms}, AFAD-lite~\cite{niu2016ordinal} and Asian dataset~\cite{zhao2018towards} for pretraining, while Feather~\cite{zhang2019feathernets} used a large private dataset with a collection protocol similar to CASIA-SURF. 2)  \emph{Many network ensembles.} VisonLab~\cite{parkin2019recognizing}, ReadSense~\cite{shen2019facebagnet}, and Feather~\cite{zhang2019feathernets} average the outputs of $24$, $12$ and $10$ networks to compute final results. Thus in order to have a fair comparison we omit VisionLab~\cite{parkin2019recognizing}, ReadSense~\cite{shen2019facebagnet} and Feather~\cite{zhang2019feathernets} from Table~\ref{tab:casia_surf_result}.

\textbf{SiW Dataset.}  Results for this dataset are shown in Table~\ref{tab:SiW}.
We compare the proposed SD-Net with other methods without pretraining. Taking the Protocol 1 of SiW as an example, SD-Net achieves the best ACER of $0.74\%$, an improvement of $0.26\%$ with respect to the second best score, $1.00\%$ (ACER) from STASN~\cite{yang2019face}. In terms of CASIA-SURF CeFA pretraining, our method is competitive to STASN (Data)~\cite{yang2019face} ($0.35\%$ versus $0.3\%$ in term of ACER), which used an large proviate dataset as pretrain.
For Protocol 2 and 3 of SiW, our methods has achieved the best performance under three evaluation metrics.

\begin{table}[t]
%\tiny
%\scriptsize
%\footnotesize
\small
\centering
\setlength{\tabcolsep}{2pt}
\scalebox{0.75}{
\begin{tabular}{|c|c|c|c|c|c|}
\hline
Prot. & Method & APCER (\%) & BPCER (\%) & ACER (\%) & Pretrain \\
\hline
\hline
\multirow{6}{*}{1} & FAS-BAS~\cite{Liu2018Learning} & 3.58 & 3.58 & 3.58 & \multirow{3}{*}{No}\\
\cline{2-5} & FAS-TD-SF~\cite{FASTD2018arxiv}       & 1.27 & \textbf{0.83} & 1.05  & \\
\cline{2-5} & STASN~\cite{yang2019face}             & - & - & 1.00  & \\
\cline{2-5} & SD-Net                                & \textbf{0.14} & 1.34 & \textbf{0.74} &  \\
\cline{2-6} & \begin{tabular}[c]{@{}c@{}}FAS-TD-SF\\ (CASIA-SURF)~\cite{DBLP:conf/cvpr/abs-1812-00408}\end{tabular}
& 1.27 & \textbf{0.33} & 0.80  & \multirow{4}{*}{Yes}\\
\cline{2-5} & STASN (Data)~\cite{yang2019face}        & - & - & \textbf{0.30}  &  \\
\cline{2-5} & SD-Net (CASIA-SURF CeFA)                           & \textbf{0.21} & 0.50 & 0.35 &  \\
\hline
\hline

\multirow{6}{*}{2} & FAS-BAS~\cite{Liu2018Learning} & 0.57$\pm$0.69 & 0.57$\pm$0.69 & 0.57$\pm$0.69 & \multirow{3}{*}{No}\\
\cline{2-5} & FAS-TD-SF~\cite{FASTD2018arxiv}       & 0.33$\pm$0.27  & \textbf{0.29$\pm$0.39} & 0.31$\pm$0.28  & \\
\cline{2-5} & STASN~\cite{yang2019face}  & - & -  & 0.28$\pm$0.05 & \\
\cline{2-5} & SD-Net & \textbf{0.25$\pm$0.32} & \textbf{0.29$\pm$0.34} & \textbf{0.27$\pm$0.28} & \\
\cline{2-6} & \begin{tabular}[c]{@{}c@{}}FAS-TD-SF\\ (CASIA-SURF)~\cite{DBLP:conf/cvpr/abs-1812-00408}\end{tabular}
& \textbf{0.08$\pm$0.17}  & 0.25$\pm$0.22 & 0.17$\pm$0.16  & \multirow{4}{*}{Yes}\\
\cline{2-5} & STASN (Data)~\cite{yang2019face}        & - & - & \textbf{0.15$\pm$0.05} & \\
\cline{2-5} & SD-Net (CASIA-SURF CeFA)                          & 0.09$\pm$0.17 & \textbf{0.21$\pm$0.25} & \textbf{0.15$\pm$0.11} & \\
\hline
\hline

\multirow{6}{*}{3} & FAS-BAS~\cite{Liu2018Learning} & 8.31$\pm$3.81  & 8.31$\pm$3.81  & 8.31$\pm$3.81 & \multirow{3}{*}{No}\\
\cline{2-5} & FAS-TD-SF~\cite{FASTD2018arxiv}       & 7.70$\pm$3.88  & \textbf{7.76$\pm$4.09}   & 7.73$\pm$3.99 & \\
\cline{2-5} & STASN~\cite{yang2019face}             & - & -  & 12.10$\pm$1.50 & \\
\cline{2-5} & SD-Net                                & \textbf{3.74$\pm$2.15} & 7.85$\pm$1.42 & \textbf{5.80$\pm$0.36} & \\

\cline{2-6} & \begin{tabular}[c]{@{}c@{}}FAS-TD-SF\\ (CASIA-SURF)~\cite{DBLP:conf/cvpr/abs-1812-00408}\end{tabular}
& 6.27$\pm$4.36  & \textbf{6.43$\pm$4.42} & 6.35$\pm$4.39  & \multirow{4}{*}{Yes}\\
\cline{2-5} & STASN (Data)~\cite{yang2019face}        & - & -  & 5.85$\pm$0.85 & \\
\cline{2-5} & SD-Net (CASIA-SURF CeFA)           & \textbf{2.70$\pm$1.56} & 7.10$\pm$1.56 & \textbf{4.90$\pm$0.00} & \\
\hline

\end{tabular}
}

\caption{Comparisons on SiW. '-' indicates unprovided; '()' means the method is used a pretrain model trained from a specific dataset. Best results are bolded in the condition of with/without pretrain. }
\label{tab:SiW}
\end{table}

\begin{table}[t]
%\tiny
%\scriptsize
%\footnotesize
\small
\centering
\setlength{\tabcolsep}{2pt}
\scalebox{0.75}{
\begin{tabular}{|c|c|c|c|c|c|}
\hline
Prot. & Method & APCER (\%) & BPCER (\%) & ACER (\%) & Pretrain\\
\hline
\hline
\iffalse
\multirow{6}{*}{1} & FAS-BAS~\cite{Liu2018Learning} & 1.6 & 1.6 & 1.6 \\
\cline{2-5} & FAS-Ds~\cite{Jourabloo2018Face}       & 1.2 & 1.7 & 1.5 \\
%\cline{2-5} & FAS-TD-SF~\cite{zhang2019casiasurf}       & 2.7 & 2.5 & 2.6 \\
\cline{2-5} & STASN~\cite{yang2019face}             & 1.2 & 2.5 & 1.9 \\
\cline{2-5} & STASN-Data~\cite{yang2019face}        & 1.2 & 0.8 & 1.0 \\
\cline{2-5} & SD-Net                                & 1.0 & 6.7 & 3.8 \\
\cline{2-5} & SD-Net (CASIA-SURF CeFA)                           & ? & ? & ? \\
\hline
\hline
\multirow{6}{*}{2} & FAS-BAS~\cite{Liu2018Learning} & 2.7 & 2.7 & 2.7 \\
\cline{2-5} & FAS-Ds~\cite{Jourabloo2018Face}       & 4.2 & 4.4 & 4.3 \\
%\cline{2-5} & FAS-TD-SF~\cite{zhang2019casiasurf}       & 2.7 & 1.6 & 2.2 \\
\cline{2-5} & STASN~\cite{yang2019face}             & 4.2 & 0.3 & 2.2 \\
\cline{2-5} & STASN-Data~\cite{yang2019face}        & 1.4 & 0.8 & 1.1 \\
\cline{2-5} & SD-Net                                & 4.3 & 1.9 & 3.1 \\
\cline{2-5} & SD-Net-CASIA-SURF CeFA                           & ? & ? & ? \\
\hline
\hline
\fi

\multirow{6}{*}{3} & FAS-BAS~\cite{Liu2018Learning} & \textbf{2.7$\pm$1.3} & 3.1$\pm$1.7 & 2.9$\pm$1.5 & \multirow{4}{*}{No}\\
\cline{2-5} & FAS-Ds~\cite{Jourabloo2018Face}       & 4.0$\pm$1.8 & 3.8$\pm$1.2 & 3.6$\pm$1.6 &  \\
%\cline{2-5} & FAS-TD-SF~\cite{zhang2019casiasurf}       & 2.4$\pm$1.5 & 2.2$\pm$3.8 & 2.3$\pm$2.6 & $\times$\\
\cline{2-5} & STASN~\cite{yang2019face}             & 4.7$\pm$3.9 & 0.9$\pm$1.2 & 2.8$\pm$1.6 & \\
\cline{2-5} & SD-Net                                & \textbf{2.7$\pm$2.5} & \textbf{1.4$\pm$2.0} & \textbf{2.1$\pm$1.4} & \\
\cline{2-6} & STASN (Data)~\cite{yang2019face}        & \textbf{1.4$\pm$1.4} & 3.6$\pm$4.6 & 2.5$\pm$2.2 & \multirow{2}{*}{Yes}\\
\cline{2-5} & SD-Net (CASIA-SURF CeFA)                           & 2.7$\pm$2.5 & \textbf{0.9$\pm$0.9} & \textbf{1.8$\pm$1.4} & \\
\hline
\hline
\multirow{6}{*}{4} & FAS-BAS~\cite{Liu2018Learning} & 9.3$\pm$5.6  & 10.4$\pm$6.0  & 9.5$\pm$6.0 & \multirow{4}{*}{No}\\
\cline{2-5} & FAS-Ds~\cite{Jourabloo2018Face}       & 5.1$\pm$6.3  & \textbf{6.1$\pm$5.1}   & 5.6$\pm$5.7 & \\
%\cline{2-5} & FAS-TD-SF~\cite{zhang2019casiasurf}       & 8.7$\pm$5.6  & 5.8$\pm$8.0   & 7.2$\pm$5.8 & $\times$\\
\cline{2-5} & STASN~\cite{yang2019face}             & 6.7$\pm$10.6 & 8.3$\pm$8.4   & 7.5$\pm$4.7 & \\
\cline{2-5} & SD-Net                                & \textbf{4.6$\pm$5.1}  & 6.3$\pm$6.3   & \textbf{5.4$\pm$2.8} & \\

\cline{2-6} & STASN (Data)~\cite{yang2019face} & \textbf{0.9$\pm$1.8}  & \textbf{4.2$\pm$5.3}   & \textbf{2.6$\pm$2.8} & \multirow{2}{*}{Yes}\\
\cline{2-5} & SD-Net (CASIA-SURF CeFA)                           & 5.0$\pm$4.7  & 4.6$\pm$4.6   & 4.8$\pm$2.7 & \\
\hline
\end{tabular}
}
\caption{Results of Protocol 3 and 4 on OULU-NPU. '()' means the method is used a pretrain model trained from a specific dataset. Best results are bolded in the conditions of with/without pretrain.}
\label{tab:Oulu}
\end{table}

\textbf{OULU-NPU Dataset.} We perform evaluation on the 2 most challenging protocols of OULU-NPU. Protocol 3 studies the generalization across different acquisition devices and Protocol 4 considers all the conditions of previous three protocols simultaneously. The experimental Results are shown in  Table~\ref{tab:Oulu}.  In the case of
comparison without pretraining, SD-Net obtains the best results in both Protocol 3 and 4. The ACER of our SD-Net is $2.1\%$ versus $2.8\%$ of STASN~\cite{yang2019face}. When comparing the results using pretraining, our method achieves the first and second position for Protocol 3 and 4, respectively. %Second, \emph{method comparison with pretrain.} Our method has get the best performance in protocol 3 while the second best performance in protocol 4.
Based on the above experiments, when without using pretraining, the proposed method can get the state-of-the-art performance (ACER) in all protocols on SiW and OULU-NPU. The proposed method with pretraining on CASIA-SURF CeFA can also get the best ACER scores on most of protocols. Those experimental results clearly demonstrate the effectiveness of the proposed method and the collected CASIA-SURF CeFA dataset.

% From STASN~\cite{yang2019face}, we know that the dataset~\cite{yang2019face} used in  STASN-Data contains more samples ($5,000$ live face videos) and attack types (over 10 attack mediums). That may be leads the more effective than our CeFA dataset.

%In the case of Protocol $4$, the result of our SD-Net is $2.1\%$ lower than STASN for ACER, and is $2.2\%$ higher when using the pre-trained dataset. The reason we analyzed is the pre-trained dataset~\cite{yang2019face} in STASN-Data contains more samples ($5,000$ live face videos) and attack types (over 10 attack mediums). However, the real samples of this dataset are selfie videos downloaded from web, which are not public available so far.

\section{Conclusion}

In this paper, we have presented the CASIA-SURF CeFA dataset for face anti-spoofing. This corresponds to the largest public available dataset in terms of modalities, subjects, ethnicities and attacks.
Moreover, we have proposed a static and dynamic network (SD-Net) to learn both static and dynamic features from single modality. Then, we have proposed a partially shared multi-modal network (PSMM-Net) to learn complementary information from multi-modal data in videos. Extensive experiments on four popular datasets show the high generalization capability of the proposed SD-Net and PSMM-Net, and the utility and challenges of the released CASIA-SURF CeFA dataset.

{\small
\bibliographystyle{ieee_fullname}
\bibliography{reference}
}

\end{document}